\documentclass[a4paper,conference]{IEEEtran}
\usepackage[T1]{fontenc}
\usepackage{color}
\usepackage{float}
\usepackage{mathrsfs}
\usepackage{amsmath}
\usepackage{amssymb}
\usepackage{graphicx}
\usepackage{float} 
\usepackage{booktabs} 
\usepackage{array} 
\usepackage{CJK}
\usepackage{indentfirst}
\usepackage{amsmath}
\usepackage{cases}
\usepackage{setspace}

 \usepackage[unicode=true,
 bookmarks=true,bookmarksnumbered=true,
 bookmarksopen=true,bookmarksopenlevel=1,
 breaklinks=false,pdfborder={0 0 0},
 pdfborderstyle={},backref=false,
 colorlinks=false]
 {hyperref}

\hypersetup{pdftitle={Your Title},
 pdfauthor={Your Name},
 pdfpagelayout=OneColumn, 
 pdfnewwindow=true, 
 pdfstartview=XYZ, 
 plainpages=false}
\makeatletter
\hypersetup{draft}

\floatstyle{ruled}
\newfloat{algorithm}{tbp}{loa}
\providecommand{\algorithmname}{Algorithm}
\floatname{algorithm}{\protect\algorithmname}

\usepackage[caption=false,font=footnotesize]{subfig}
\usepackage{algorithm}
\usepackage{algorithmic}

\usepackage{multirow} 
\usepackage{amsmath} 
\usepackage{xcolor}

\allowdisplaybreaks[4]

\ifCLASSOPTIONcompsoc
\usepackage[caption=false,font=normalsize,labelfont=sf,
textfont=sf]{subfig}
\else
\usepackage[caption=false,font=footnotesize]{subfig}
\fi

\usepackage{cite}
\usepackage{bm}
\usepackage{algorithmic}
\usepackage{algorithm}
\usepackage{graphicx}
\IEEEoverridecommandlockouts

\usepackage{lettrine}

\usepackage{geometry}
\usepackage{subfig}
\makeatother

\begin{document}

\title{\textcolor{black}{Hierarchical Task Offloading for UAV-Assisted Vehicular Edge Computing via Deep Reinforcement Learning}}
\author{
\IEEEauthorblockN{Hongbao Li$^{\dagger}$, 
Ziye Jia$^{\dagger}$,
Sijie He$^{\dagger}$, Kun Guo$^{ \star }$, and Qihui Wu$^{\dagger}$  \\}
\IEEEauthorblockA{$^{\dagger}$The Key Laboratory of Dynamic Cognitive System of Electromagnetic Spectrum Space, Ministry of Industry and\\ 
Information Technology, Nanjing University of Aeronautics and Astronautics, Nanjing, Jiangsu, 211106, China\\
$^{ \star }$School of Communications and Electronics Engineering, East China Normal University, Shanghai, 200241, China\\
 lihongbao491@gmail.com, \{jiaziye, hesijie, wuqihui\}@nuaa.edu.cn, kguo@cee.ecnu.edu.cn}

 \thanks{{This work was supported in part by National Natural Science Foundation of China under Grant 62301251, in part by  the open research fund of National Mobile Communications Research Laboratory, Southeast University (No. 2024D04), in part by the Aeronautical Science Foundation of China 2023Z071052007, and in part by the Young Elite Scientists Sponsorship Program by CAST 2023QNRC001 (Corresponding author: Ziye Jia).
 }} }

\maketitle

\pagestyle{empty} 

\thispagestyle{empty}

\begin{spacing}{1}
\begin{abstract}
    With the emergence of compute-intensive and delay-sensitive applications in vehicular networks, unmanned aerial vehicles (UAVs) have emerged as a promising complement for vehicular edge computing due to the high mobility and flexible deployment. However, the existing UAV-assisted offloading strategies are insufficient in coordinating heterogeneous computing resources and adapting to dynamic network conditions. Hence, this paper proposes a dual-layer UAV-assisted edge computing architecture based on partial offloading, composed of the relay capability of high-altitude UAVs and the computing support of low-altitude UAVs. The proposed architecture enables efficient integration and coordination of heterogeneous resources. A joint optimization problem is formulated to minimize the system delay and energy consumption while ensuring the task completion rate. To solve the high-dimensional decision problem, we reformulate the problem as a Markov decision process and propose a hierarchical offloading scheme based on the soft actor-critic algorithm. The method decouples global and local decisions, where the global decisions integrate offloading ratios and trajectory planning into continuous actions, while the local scheduling is handled via designing a priority-based mechanism. Simulations are conducted and demonstrate that the proposed approach outperforms several baselines in task completion rate, system efficiency, and convergence speed, showing strong robustness and applicability in dynamic vehicular environments.

\begin{IEEEkeywords}
    UAV-assisted vehicular networks, computation offloading, deep reinforcement learning, priority-aware resource scheduling.
\end{IEEEkeywords}
\end{abstract}
\end{spacing}
\newcommand{\CLASSINPUTtoptextmargin}{0.8in}

\newcommand{\CLASSINPUTbottomtextmargin}{1in}

\section{Introduction}
\begin{spacing}{1}
    \lettrine[lines=2, lhang=0.05, loversize=0.15]{W}{ith} the rapid development of intelligent transportation systems, vehicular terminals have been widely deployed in advanced applications, such as autonomous driving, high-precision environmental perception, path planning, and in-vehicle entertainment{\cite{hu2020regional}}. These services involve a large number of compute-intensive and delay-sensitive tasks, bringing increasingly stringent demands on onboard computational capabilities. The traditional resource allocation mechanisms are insufficient for growing requirements of timely and reliable task processing{\cite{zhang2018vehicular}}.
    
    Generally, the vehicular edge computing (VEC) frameworks primarily rely on static roadside units (RSUs) or base stations (BSs) to offload and process tasks{\cite{You2025CISS}}. While the architectures may alleviate onboard computational load and reduce response delay, they are constrained by limited coverage and inflexible deployment locations{\cite{liu2023uav}}. These limitations make them insufficient to cope with rapidly changing vehicular topology and unbalanced regional load distributions{\cite{hu2021uav}}. Due to the high mobility, unmanned aerial vehicles (UAVs) have been increasingly adopted in VEC systems as a powerful supplement for enhancing the task offloading flexibility and network adaptability{\cite{Jia2024VTM,10638237,jia2025dro}}.
    
    However, the existing UAV-assisted offloading strategies  face challenges in practical deployments. Firstly, many studies fail to effectively integrate heterogeneous aerial-ground resources and lack path selection mechanisms tailored to task heterogeneities{\cite{li2023flexedge,yu2020joint,tun2021energy,jia2025sfc}}. In particular, most works overlook the opportunity to relay non-delay-sensitive tasks to remote idle RSUs or BSs to alleviate the hotspot congestion, resulting in underutilization of global resources. Besides, RSUs are prone to overload in high-traffic-density scenarios, without dynamic diversion mechanisms, regional computational bottlenecks may occur{\cite{dai2024uav}}. Moreover, traditional optimization methods exhibit poor adaptabilities  to dynamic vehicular environments with rapid topology evolution, while direct applications of deep reinforcement learning (DRL) face the curse of dimensionality in the action spaces and training instability, which hinder their practical applicabilities {\cite{9026875}}.
    
    In this work, we propose a dual-layer UAV-assisted edge computing architecture based on partial-offloading. The proposed system integrates the high-altitude UAV (HUAV) with global relaying and coordination ability and the low-altitude UAV (LUAV) with localized deployment and rapid responsiveness ability, forming an efficient aerial-ground collaborative task processing system. To optimize the collaboration, we formulate a joint optimization problem that aims to minimize the overall system delay and energy consumption while maintaining a high task completion rate. The model jointly considers multiple decision variables including the task offloading ratio, offloading node selection, computing resource allocation, and LUAV trajectory planning. To solve the complex optimization problem, it is reformulated as a Markov decision process (MDP), and a soft actor-critic (SAC)-based hierarchical offloading algorithm is proposed. The algorithm decouples the action space, where the DRL agent is responsible for determining the global offloading ratios and LUAV trajectories, while node selection and computing resource scheduling are handled by a priority-based mechanism. Simulation results demonstrate that the proposed approach outperforms baselines in task completion rate and system efficiency, exhibiting strong robustness in dynamic vehicular environments. 

    
        
    

\textcolor{black}{
The reminder of the paper is organized as follows. Section~\ref{sec:System-Model} introduces the system model and problem formulation. The SAC-based offloading algorithm is presented in Section~\ref{sec:Reinforcement-learning}. Section~\ref{sec:SimulationResult} provides simulation results and performance comparisons. Section~\ref{sec:Conclusions} concludes the paper.
}
\end{spacing}

\section{System Model and Problem Formulation\label{sec:System-Model}}

\subsection{Network Model}
\begin{figure}[t]
    \centering
    \includegraphics[width=0.85\linewidth]{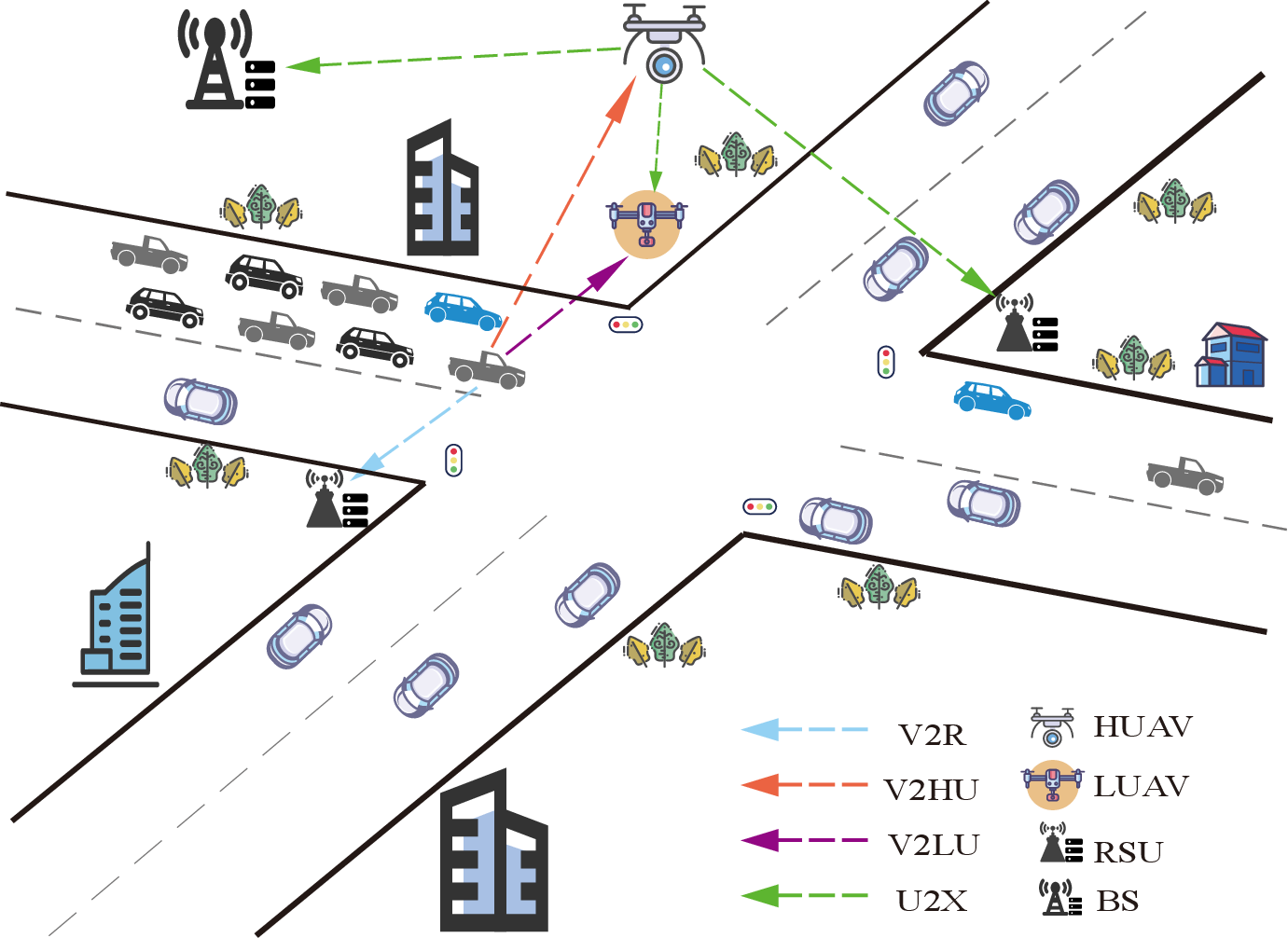}
    \caption{System model of the dual-layer UAV-assisted vehicular edge computing network. The architecture includes vehicles, RSUs, one HUAV, and multiple LUAVs. For clarity, only one LUAV is shown.}
    \label{fig:SystemModel}
\end{figure}

\begin{spacing}{0.98}
We consider a typical urban vehicular network scenario in Fig.~\ref{fig:SystemModel}. The network consists of $I$ intelligent vehicles, $R$ RSUs, $L$ LUAVs, one HUAV, and one high-performance BS. To facilitate analysis, we denote the sets of intelligent vehicles, RSUs, and LUAVs as $\mathcal{I} = \{1, \dots,i,\dots, I\}$, $\mathcal{R} = \{1, \dots,r,\dots, R\}$, and $\mathcal{L} = \{1, \dots,l,\dots, L\}$, respectively. The HUAV and BS are represented by $H$ and $B$, respectively. We define the set of edge computing nodes as \( x \in \{\mathcal{L} \cup \mathcal{R} \cup B \}\). Each node \( x \) is equipped with limited computational resources denoted by \( F_{x}^{\max} \). The entire duration $T$ is divided into $N$ equal-length time slots, indexed by $\mathcal{T} = \{1, \dots, t, \dots, N\}$. Within each slot $t \in \mathcal{T}$, the node positions are assumed to be static{\cite{Diao2021UAV}}. At the beginning of each slot, RSUs report the regional traffic to the HUAV, which then predicts the traffic hotspots and dispatches LUAVs. Each vehicle $i \in \mathcal{I}$ is located at $\mathbf{p}_{i} = (x_{i}, y_{i})$ and moves at velocity of $v_{i}$. The RSUs and BS are deployed at fixed locations denoted by $\mathbf{p}_r$ and $\mathbf{p}_B$, respectively.  The HUAV hovers at position $\mathbf{p}_H = (x_H, y_H, z_H)$. Each LUAV $l$ is located at $\mathbf{p}_l = (x_l, y_l, z_{l})$, where $z_{l}$ denotes a fixed flight altitude. The LUAV moves horizontally with direction $q_l$ and speed $v_l$, limited by a maximum speed constraint $v_l^{\max}$. Due to the altitude differences, the vehicles, RSUs, and BSs are modeled as ground-level nodes.
\end{spacing}

\subsection{Task Generation Model}
In each time slot $t \in \mathcal{T}$, every vehicle $i \in \mathcal{I}$ generates a computation task, represented as
\[
\tau_{i}(t) = \left(D_{i}(t),\; C_{i}(t),\; T^{\max}_{i}(t),\; K_{i}(t) \right),
\]
where $D_{i}(t)$ is the data size, $C_{i}(t)$ is the required CPU cycles, $T^{\max}_{i}(t)$ is the maximum tolerable delay, and $K_{i}(t)$ denotes the task priority. This priority is jointly determined by the resource intensity and delay
sensitivity of the task, with larger values indicating higher priorities.

\subsection{Communication Model}

\subsubsection{Vehicle-to-RSU Communication}

In urban environments, due to building obstructions, the vehicle-to-RSU communication follows a non-line-of-sight (NLoS) ground-to-ground (G2G) model{\cite{10616106,wang2024sparse}}. Considering the log-normal shadow fading $\xi \sim \mathrm{Log\text{-}Normal}(\mu,\sigma^2)$, the channel gain $h_{i,r}$ between vehicle $i$ and RSU $r$ is
\begin{align}
h_{i,r} &= \frac{\beta_0 \xi}{d_{i,r}^{\alpha_1}}, \label{eq:rsu_channel_gain}
\end{align}
where $\beta_0$ is the reference channel gain at 1 meter, $\alpha_1$ is the path loss exponent of NLoS channel. $d_{i,r}$ is the Euclidean distance between vehicle $i$ and RSU $r$. Further, the achievable uplink rate is
\begin{align}
R_{i,r} &= B_{i,r} \log_2\left(1 + \frac{P_{i}   h_{i,r}}{P_n} \right), \label{eq:rsu_rate}
\end{align}
where $B_{i,r}$ is the allocated bandwidth, $P_{i}$ is the transmit power of the vehicle $i$, and $P_n$ is the noise power.

\subsubsection{Vehicle-to-LUAV Communication}

We assume the line-of-sight (LoS) communication between the vehicles and LUAV due to short distances and low-altitude hovering. Hence, the channel gain is
\begin{align}
h_{i,l} &= \frac{\beta_0}{d_{i,l}^{\alpha_2}}, \label{eq:luav_channel_gain}
\end{align}
where $\alpha_2$ is the path loss exponent of LoS channel, $d_{i,l}$ is the Euclidean distance between vehicle $i$ and LUAV $l$. Hence, the corresponding data rate is
\begin{align}
R_{i,l} &= B_{i,l} \log_2\left(1 + \frac{P_{i}   h_{i,l}}{P_n} \right), \label{eq:luav_rate}
\end{align}
in which $B_{i,l}$ is the bandwidth allocated to vehicle $i$ by LUAV $l$.

\subsubsection{Vehicle-to-HUAV-to-RSU/BS Communication}

The HUAV maintains LoS communication links with both vehicles and idle RSUs or the BS. The corresponding channel gains are modeled as
\begin{align}
    h_{i,H} &= \frac{\beta_0}{d_{i,H}^{\alpha_2}}, \label{eq:huav_channel_gain_1} \\
    \intertext{and}
    h_{H,x} &= \frac{\beta_0}{d_{H,x}^{\alpha_2}}, \quad x \in \{\mathcal{R} \cup B\}. \label{eq:huav_channel_gain_2}
    \end{align}
Wherein, $d_{i,H}$ and $d_{H,x}$ denote the Euclidean distances from vehicle $i$ to the HUAV and from the HUAV to node $x$, respectively. Hence, the data rates of the two-hop relay path are
\begin{align}
R_{i,H} &= B_{i,H} \log_2\left(1 + \frac{P_{i}   h_{i,H}}{P_n} \right), \label{eq:huav_rate_1} \\
\intertext{and}
 R_{H,x} &= B_{H,x} \log_2\left(1 + \frac{P_{H}   h_{H,x}}{P_n} \right), \label{eq:huav_rate_2}
\end{align}
where $P_{H}$ is the transmit power of the HUAV. $B_{i,H}$ and $B_{H,x}$ are the corresponding bandwidths for each hop, respectively.

\subsection{Task Offloading Model}

In the considered vehicular edge computing system, each task $\tau_i(t)$ at time slot $t$ can be partially processed locally and concurrently offloaded to edge computing nodes.
Let $\mathcal{K} = \{\text{LOCAL}, \text{RSU}, \text{LUAV}, \text{HUAV-RSU}, \text{HUAV-BS}\}$ denote the set of offloading modes. The offloading ratio along mode $k \in \mathcal{K}$ for vehicle $i$ at time slot $t$ is denoted by $\lambda_{i}^{k}(t) \in [0,1]$, subject to the constraint $\sum_{k \in \mathcal{K}} \lambda_{i}^{k}(t) = 1$. The delay and energy consumption under each offloading mode are modeled as follows.

\subsubsection{Local Computation}

When the task is processed locally on the vehicle ($k = \text{LOCAL}$), the execution delay depends on the CPU frequency $f_{i}(t)$ of vehicle $i$, i.e.,
\begin{equation}
T_{i}^{k}(t) = \frac{\lambda_{i}^{k}(t)   C_{i}(t)}{f_{i}(t)}.
\end{equation}

The corresponding energy consumption $E_{i}^{k}(t)$ follows a cubic relationship with respect to the CPU frequency, i.e.,
\begin{equation}
E_{i}^{k}(t) = \eta_{i}   \left(f_{i}(t)\right)^3   T_{i}^{k}(t),
\end{equation}
where $\eta_{i}$ is the effective switched capacitance of vehicle $i$.

\subsubsection{Offloading to LUAV/RSU}

If the task is offloaded to a nearby LUAV or RSU (i.e., \( k = \text{LUAV} \) or \( k = \text{RSU} \)), the corresponding edge node is denoted by $x \in \{\mathcal{L} \cup \mathcal{R}\}$. The total delay includes the data transmission time from vehicle $i$ to edge node $x$, and the execution time at the node $x$, i.e.,
\begin{equation}
T_{i}^{k}(t) = \frac{\lambda_{i}^{k}(t)   D_{i}(t)}{R_{i,x}} + \frac{\lambda_{i}^{k}(t)   C_{i}(t)}{f_{x}^{i}(t)},
\end{equation}
where $f_{x}^{i}(t)$ denotes the CPU frequency allocated to $\tau_{i}(t)$ at node $x$.

The corresponding energy consumption consists of the transmission energy of vehicle and the computing energy, i.e.,
\begin{equation}
E_{i}^{k}(t) = P_{i}   \frac{\lambda_{i}^{k}(t)   D_{i}(t)}{R_{i,x}} + \eta_x   \left(f_{x}^{i}(t)\right)^3   \frac{\lambda_{i}^{k}(t)   C_{i}(t)}{f_{x}^{i}(t)},
\end{equation}
where $\eta_x$ is the effective switched capacitance of node $x$.

As for LUAVs, the propulsion energy during each time slot depends on the LUAV velocity, i.e.,
\begin{equation}
\label{eq:propulsion_power}
\begin{aligned}
P(v_l) &= P_o^L \left(1 + \frac{3v_l^2}{U_{\text{tip}}^2} \right) + \frac{1}{2} d_0 \rho_s A v_l^3 \\
&\quad + P_i^L \left( \sqrt{1 + \frac{v_l^4}{4\nu_0^4}} - \frac{v_l^2}{2\nu_0^2} \right)^{1/2},
\end{aligned}
\end{equation}
where $U_{\text{tip}}$, $A$, and $\rho$ denote the rotor tip speed, rotor disk area, and air density, respectively. The parameters $d_0$, $\nu_0$, and $s$ correspond to the body drag coefficient, induced velocity, and blade area ratio, respectively. $P_o^L$ and $P_i^L$ represent the profile and induced power of the LUAV, respectively.

Thus, the energy consumption of LUAV $l$ for flying during time slot $t$ is
\begin{equation}
E_{l}^{\text{fly}}(t) = P(v_l)   \frac{T}{N}.
\end{equation}

\subsubsection{Relay via the HUAV}
For the delay-tolerant tasks, offloading can be performed via a HUAV to alleviate congestions on local edge nodes{\cite{9714482}}. Specifically, tasks are relayed to an idle RSU or the BS (i.e., \( k = \text{HUAV-RSU} \) or \( k = \text{HUAV-BS} \)). The corresponding target node $x$ belongs to the set $\{(\mathcal{R} \setminus \{r_i^{\mathrm{direct}}\}) \cup B\}$, where $r_i^{\mathrm{direct}}$ denotes the RSU directly connected to vehicle $i$. The total delay  consists of the two-hop transmission delay and computation delay at remote node $x$, i.e.,
\begin{multline}
T_{i}^{k}(t) = \frac{\lambda_{i}^{k}(t)   D_{i}(t)}{R_{i,H}} 
+ \frac{\lambda_{i}^{k}(t)   D_{i}(t)}{R_{H,x}} 
 + \frac{\lambda_{i}^{k}(t)   C_{i}(t)}{f_{x}^{i}(t)}.
\end{multline}

The energy consumption for  offoading relayed by the HUAV is composed of the transmission energy of vehicle $i$ and HUAV, as well as the computation energy at remote node $x$, i.e.,
\begin{multline}
E_{i}^{k}(t) = P_{i}   \frac{\lambda_{i}^{k}(t)   D_{i}(t)}{R_{i,H}} 
+ P_{H}   \frac{\lambda_{i}^{k}(t)   D_{i}(t)}{R_{H,x}} \\
+ \eta_x   \left(f_{x}^{i}(t)\right)^3   \frac{\lambda_{i}^{k}(t)   C_{i}(t)}{f_{i}(t)^{x}}.
\end{multline}

Finally, since the HUAV remains hovering during data relaying and coordination, the hovering energy consumption per time slot is
\begin{equation}
E_{H}^{\text{hover}}(t) = \left(P_o^H + P_i^H\right)   \frac{T}{N},
\end{equation}
where $P_o^H$ and $P_i^H$ are the profile power and induced power of the HUAV, respectively.

\subsection{Problem Formulation}
Since the task can be computed in parallel, the total delay of task \( T_{i}^{\text{total}}(t) \) is the maximum delay among all offloading modes, i.e.,
\begin{equation}
    T_{i}^{\text{total}}(t) = \max_{k \in \mathcal{K}} T_{i}^{k}(t).
    \label{eq:total_delay}
    \end{equation}
The task completion rate is defined as
\begin{equation}
R^{\text{succ}} = \frac{1}{|I|} \sum_{i \in \mathcal{I}} 1\left\{ T_{i}^{\text{total}}(t) \leq T_{i}^{\text{max}}(t) \right\},
\end{equation}
where \( 1\{ \} \) is the indicator function, which returns 1 if the condition inside the brackets is true and 0 otherwise.

The energy consumption for a single task includes the computation energy consumed by offloading to each mode:
\begin{equation}
    E_{i}^{\text{comp}}(t) = \sum_{k \in \mathcal{K}} E_{i}^{k}(t).
    \end{equation}
Additionally, the total UAV energy consumption in time slot $t$ is
\begin{equation}
E^{\text{UAV}}(t) = \sum_{l=1}^{L} E_l^{\text{fly}}(t) + E_{H}^{\text{hover}}(t).
\end{equation}
Thus, the total energy consumption of the system in time slot $t$ is
\begin{equation}
E^{\text{sys}}(t) = \sum_{i \in \mathcal{I}} E_{i}^{\text{comp}}(t) + E^{\text{UAV}}(t).
\end{equation}

Based on the above delay and energy metrics, the optimization objective is to minimize the total system delay and energy consumption while ensuring the quality of service requirements, i.e.,
\begin{subequations} \label{optimal}
    \begin{align}
    \text{P0}:\underset{\boldsymbol{\lambda}, \mathbf{F}, \mathbf{Q}}{\max} & = \omega_1 R^{\text{succ}} - \omega_2 \beta_T \sum_{i \in \mathcal{I}} T_{i}^{\text{total}}(t) - \omega_3 \beta_E E^{\text{sys}}(t)  
    \\
    \text{s.t.}\quad
    & \sum_{k \in \mathcal{K}} \lambda_{i}^{k}(t) = 1,  \quad \forall i \in \mathcal{I}, 
    \\
    & \sum_{i \in \mathcal{I}_x(t)} f_x^i(t) \leq F_{x}^{\max}, \quad \forall x \in \{\mathcal{L} \cup \mathcal{R} \cup B\}
    \\
    & \|\mathbf{p}_{l}(t+1) - \mathbf{p}_{l}(t)\| \leq v_{lu_{max}}   \frac{T}{N}, \quad \forall l \in \mathcal{L},\\
    & \sum_{t=1}^{N} E_l^{\text{fly}}(t) \leq E_{l}^{\text{max}},\quad \forall l \in \mathcal{L},\\
    & \sum_{t=1}^{N} E_{H}^{\text{hover}}(t) \leq E_{H}^{\text{max}},\\
    & \lambda_{i}^k(t) \geq 0, \quad \forall i \in \mathcal{I},\ \forall k \in \mathcal{K}.
    \end{align}
    \end{subequations}
Wherein, $\omega_1,\, \omega_2,\, \omega_3 \in [0,1]$ are weight parameters and \( \omega_1 + \omega_2 + \omega_3 = 1 \). \( \beta_T \) and \( \beta_E \) are normalization parameters for delay and energy consumption, respectively. \( \bm{\lambda} = \{ \lambda_{i}^k(t) \}_{i \in \mathcal{I},k \in \mathcal{K}} \) represents the offloading decision matrix, where each entry specifies the proportion of the task $\tau_i(t)$ assigned to mode \( k \) at time slot \( t \).   \( \mathbf{F} = \{ f_{x}^i(t) \}_{x \in \{\mathcal{L} \cup \mathcal{R} \cup B\}, i \in \mathcal{I}} \) is the computation resource allocation variable representing the resources assigned by node \( x \) to the  vehicle $i$ at time slot \( t \). \( \mathbf{Q} = \{ q_{l}(t), v_{l}(t) \}_{l \in \mathcal{L}} \) denote the set of trajectory variables for all LUAVs, where \( q_{l}(t) \in [0, 2\pi] \) and \(  v_{l}(t) \in [0, v_{l}^{\max}] \).

Note that P0 is a mixed-integer nonlinear programming problem, as it involves both discrete offloading decisions and continuous resource allocation and trajectory variables. Additionally, the dynamic positions of vehicles and LUAVs, coupled with the time-varying task demands, further exacerbate the computational complexity.  Such dynamic and nonlinear characteristics pose substantial challenges for conventional optimization methods.

\section{Hierarchical SAC-Based Offloading Algorithm\label{sec:Reinforcement-learning}}
\begin{figure}[t]
    \begin{minipage}{0.48\textwidth}
    \includegraphics[width=\linewidth]{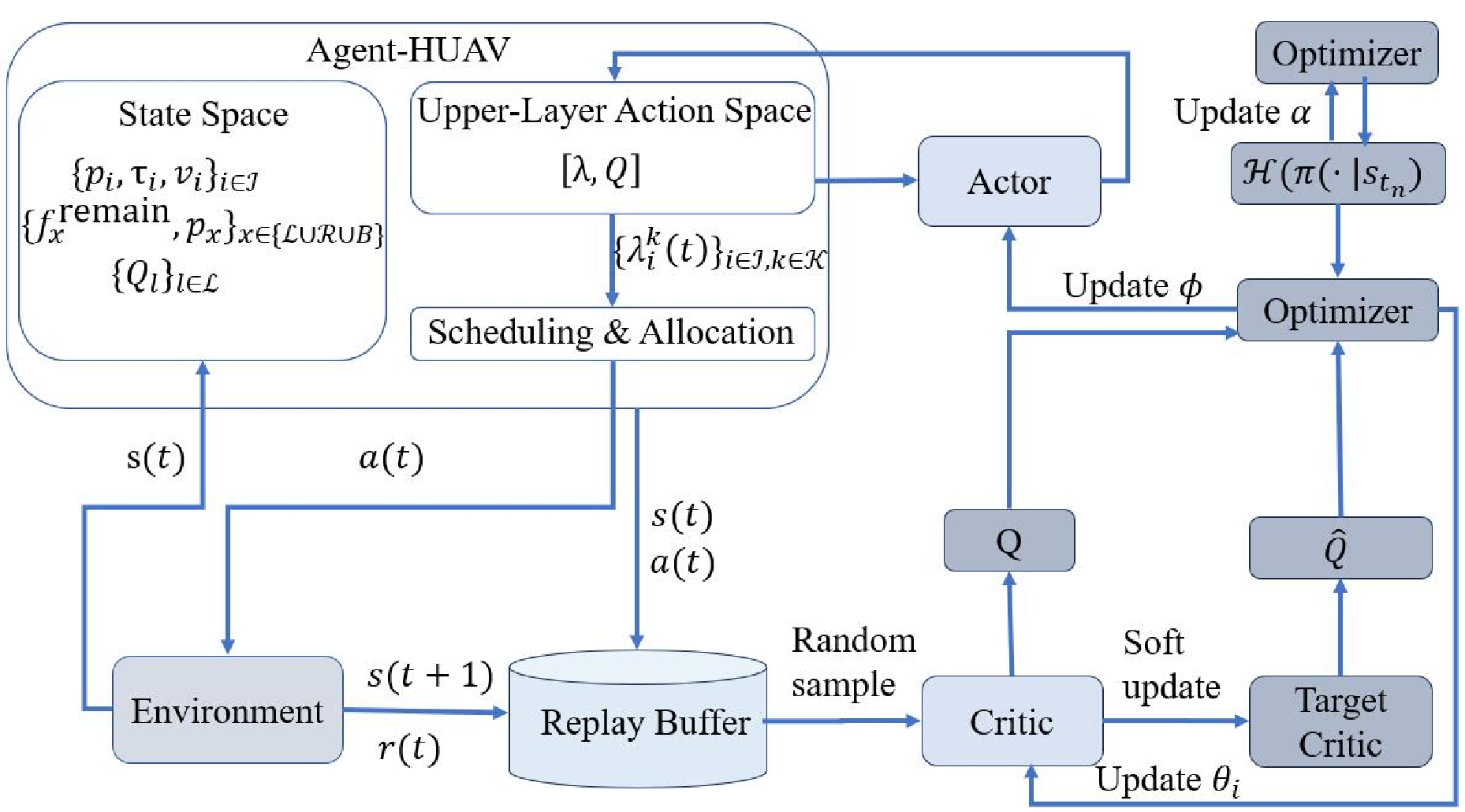}
    \caption{The architecture of the proposed hierarchical offloading algorithm.}
    \label{fig:algorithm_architecture}
\end{minipage}
\end{figure}

To address the aforementioned challenges and enable efficient task offloading and resource allocation in multi-source vehicle edge computing, we propose a hierarchical SAC-based offloading strategy deployed on the HUAV. The strategy consists of a policy decision layer and a resource scheduling layer. In detail, the policy layer leverages the SAC algorithm to determine offloading ratios and LUAV trajectories, while the scheduling layer adopts the priority-based logic for node selection and resource allocation. The structure is illustrated in Fig.~\ref{fig:algorithm_architecture}.

\subsection{Reinforcement Learning Modeling and SAC Algorithm Design}
To jointly optimize the task offloading and UAV trajectory under dynamic conditions, we reformulate the problem as an MDP. In detail, the HUAV serves as the intelligent agent, observing the global state and issuing scheduling actions at each time slot \( t_n \). The process is defined as \( (S, A, P, R, \gamma) \).

\subsubsection{State Space}
The state at time slot \( t \) is
\begin{equation}
\begin{aligned}
s(t) = \big\{ 
    \{ \mathbf{p}_{i}, \tau_{i}, v_i \}_{i \in \mathcal{I}}, \{ f_x^{\text{remain}}, \mathbf{p}_x \}_{x \in \{\mathcal{L} \cup \mathcal{R} \cup B\}}, \{ Q_l \}_{l \in \mathcal{L}} 
\big\},
\end{aligned}
\end{equation}
where $\{ \mathbf{p}_{i}, \tau_{i}, v_i \}$ represents the basic status of vehicle $i$. \( f_x^{remain} \) and \(\mathbf{p}_x \) denote the node resource and location, respectively. \( Q_{l} \) denotes the LUAV motion state. The state includes  the full network status for decision-making.

\subsubsection{Action Space}
The action vector $a(t)$ includes offloading ratios and LUAV trajectory control, i.e.,
\begin{equation}
a(t) = \{ \bm{\lambda}, \mathbf{Q} \}.
\end{equation}

\subsubsection{Reward Design}
The immediate reward $r(t)$ encourages high task completion with low delay and energy
\begin{equation}
r(t) = \omega_1 R^{\text{succ}}(t) - \omega_2 \beta_T T^{\text{total}}(t) - \omega_3\beta_E E^{\text{sys}}(t),
\end{equation}
where \( T^{\text{total}} = \sum_{i \in \mathcal{I}} T_{i}^{\text{total}}(t)  \) is the system delay.
Further, we design the SAC algorithm to realize upper-layer task offloading decisions. The objective is formulated as

\begin{equation}
\pi^* = \arg \max_\pi \mathbb{E} \left[ \sum_t \left( r(s_t, a_t) + \alpha H(\pi(  | s_t)) \right) \right],
\end{equation}
where \( r(s_t, a_t) \) is the immediate reward, \( H(\pi(  | s_t)) \) denotes the policy entropy, and \( \alpha \) is the temperature coefficient controlling the exploration-exploitation trade-off. By encouraging stochasticity in decision-making, SAC enables stable and efficient learning in dynamic environments. 
\subsection{Task Scheduling and Resource Allocation}

To improve the overall task completion rate and ensure timely responses for critical tasks, a priority-based scheduling strategy is adopted. At each time slot, tasks are sorted in an ascending order based on task type \( K_{i}(t) \) and maximum tolerable delay \( T_{i}^{max}(t) \), allowing high-priority and delay-sensitive tasks to be scheduled first.

For non-local offloading modes, the system determines the target offloading node based on two basic criteria: (1) the communication link must be reachable, and (2) the node must have more than 30\% of its computational resources available.

The candidate nodes are then scored based on both the communication distance and remaining computational resources of the nodes

\begin{equation}
\text{Score}_{i,x} = \frac{\alpha_s}{d_{i,x}} + \beta_s   \frac{f_x^{\text{remain}}}{F_x^{\max}},
\end{equation}
where \( d_{i,x} \) is the Euclidean distance between vehicle \( i \) and node \( x \), \( f_x^{\text{remain}} \) denotes the remaining computational resources of node \( x \). \( \alpha_s \) and \( \beta_s \) are tunable weighting parameters. The system prioritizes nodes with higher scores and allocates resources to tasks. If there exists no suitable node for a given mode, it falls back to the next best mode until all tasks are processed.

The overall training and execution process of the proposed hierarchical offloading framework is summarized in Algorithm~\ref{alg:hierarchical_offloading}. Specifically, the training process begins by initializing the network environment, including node resources and communication settings. At the start of each episode, the environment is reset and the initial state $s_0$ is obtained. Based on the observed state, the SAC agent generates global actions, including offloading ratios and LUAV trajectories. These actions are then used to guide the task scheduling and computing resource allocation through the priority-based mechanism. By interacting with the environment, the system obtains the next state and corresponding reward, and stores the resulting transition in the replay buffer. The SAC networks are periodically updated using the mini-batches sampled from the buffer. This process continues until the predefined maximum time slot $T$ is reached, marking the end of the episode. The overall training process terminates when the predefined number of simulation episodes $T_{sim}$ is reached.

\begin{algorithm}[t]
    \caption{Hierarchical SAC-Based Offloading Algorithm}
    \label{alg:hierarchical_offloading}
    \begin{algorithmic}[1]
    \REQUIRE Number of vehicles $N_{\text{veh}}$, LUAV count, max training episodes $T_{sim}$, time slots per episode $T$, policy update frequency $f_{\text{update}}$.
    \ENSURE Trained policy $\pi^*$.
    
    \STATE Initialize network topology, communication channels, node resources, SAC agent and replay buffer.
    \FOR{each episode $= 1$ to $T_{sim}$}
        \STATE Reset environment and obtain initial state $s_0$.
        \FOR{each time slot $t = 1$ to $T$}
            \STATE Select offloading ratios and LUAV trajectory via SAC agent.
            \STATE Execute priority-based scheduling and resource allocation.
            \STATE Interact with environment and get $s_{t+1}$ and $r_t$.
            \STATE Store transition $(s_t, a_t, r_t, s_{t+1})$ in replay buffer.
            \IF{$t \bmod f_{\text{update}} = 0$}
                \STATE Sample mini-batch from buffer and update SAC networks.
            \ENDIF
            \STATE $s_t \leftarrow s_{t+1}$.
        \ENDFOR
        \STATE Log performance metrics for the episode.
    \ENDFOR
    \RETURN Optimized policy $\pi^*$.
    \end{algorithmic}
    \end{algorithm}

\addtolength{\topmargin}{0.095in}

\section{Simulation Results \label{sec:SimulationResult}}
To validate the effectiveness of the proposed algorithm, we conduct numerical experiments and compare its performance with three benchmark algorithms:
\begin{itemize}
\item \textbf{NoPriority}: Removes the priority scheduling module from the proposed method, relying on a FIFO queue that ignores vehicle demand heterogeneity.
\item \textbf{FixedUAV}: Disables HUAV trajectory optimization, preventing dynamic adjustment of LUAV trajectories based on traffic conditions.
\item \textbf{DQN-based}: Employs a traditional deep Q-network with discrete action space, limiting decision granularity compared to the SAC-based method.
\end{itemize}

The simulations are implemented on the PyCharm platform using Python 3.8 and PyTorch under a Windows environment. The vehicles, UAVs, RSUs, and BS are deployed over a $2,000\mathrm{m} \times 2,000\mathrm{m}$ urban area following an intersection-based topology. The channel gain at reference distance is set as $\beta_0 = -50$ dB, with path loss exponents for NLoS and LoS links as $\alpha_1 = 3.5$ and $\alpha_2 = 2$, respectively. The noise power is fixed at $P_n = -110$ dBm. Vehicles have a speed range of $v_i \in [0, 20]$ m/s, transmit power $P_i = 0.1$ W, and maximum CPU frequency $F_i^{\max} = 0.5$ GHz. RSUs and BSs are equipped with CPU frequencies $F_r^{\max} = 8$ GHz and $F_B^{\max} = 10$ GHz. The LUAV flies at an altitude of $z_l = 20$ m with CPU frequency $F_l^{\max} = 5$ GHz, while the HUAV flies at $z_H = 100$ m. Task data sizes range from $D_i \in [100, 600]$ Mbit, with maximum tolerable delay $T_i^{\max} \in [0.1, 1]$ seconds. To evaluate the overall system performance, the system utility is defined in accordance with the reward function, with the weights set as $\omega_1 = 0.6$, $\omega_2 = 0.2$, and $\omega_3 = 0.2$, respectively.

\begin{figure}[t]
    \begin{minipage}{0.48\textwidth}
    \centering
    \includegraphics[width=0.7\linewidth]{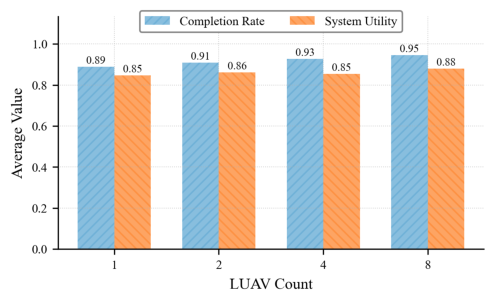}
    \caption{System utility versus episode under different LUAV numbers (Vehicles = 20, HUAV bandwidth = 100\,MHz).}
    \label{fig:system_utility_luav}
\end{minipage}
\end{figure}

Fig.~\ref{fig:system_utility_luav} presents the task completion rate and system utility versus episodes under different LUAV deployments. It can be observed that increasing the number of LUAVs significantly enhances the task completion rate. However, while the completion rate improves from LUAV = 2 to LUAV = 4, the system utility has no increase due to the additional energy consumption incurred by deploying more UAVs. As the number of LUAVs increases to 8, both the completion rate and the system utility reach the highest values. The results highlight the need to balance the task offloading performance and energy efficiency when determining the number of LUAVs deployed in the network.

\begin{figure}[t]
    \begin{minipage}{0.48\textwidth}
    \centering
    \includegraphics[width=0.7\linewidth]{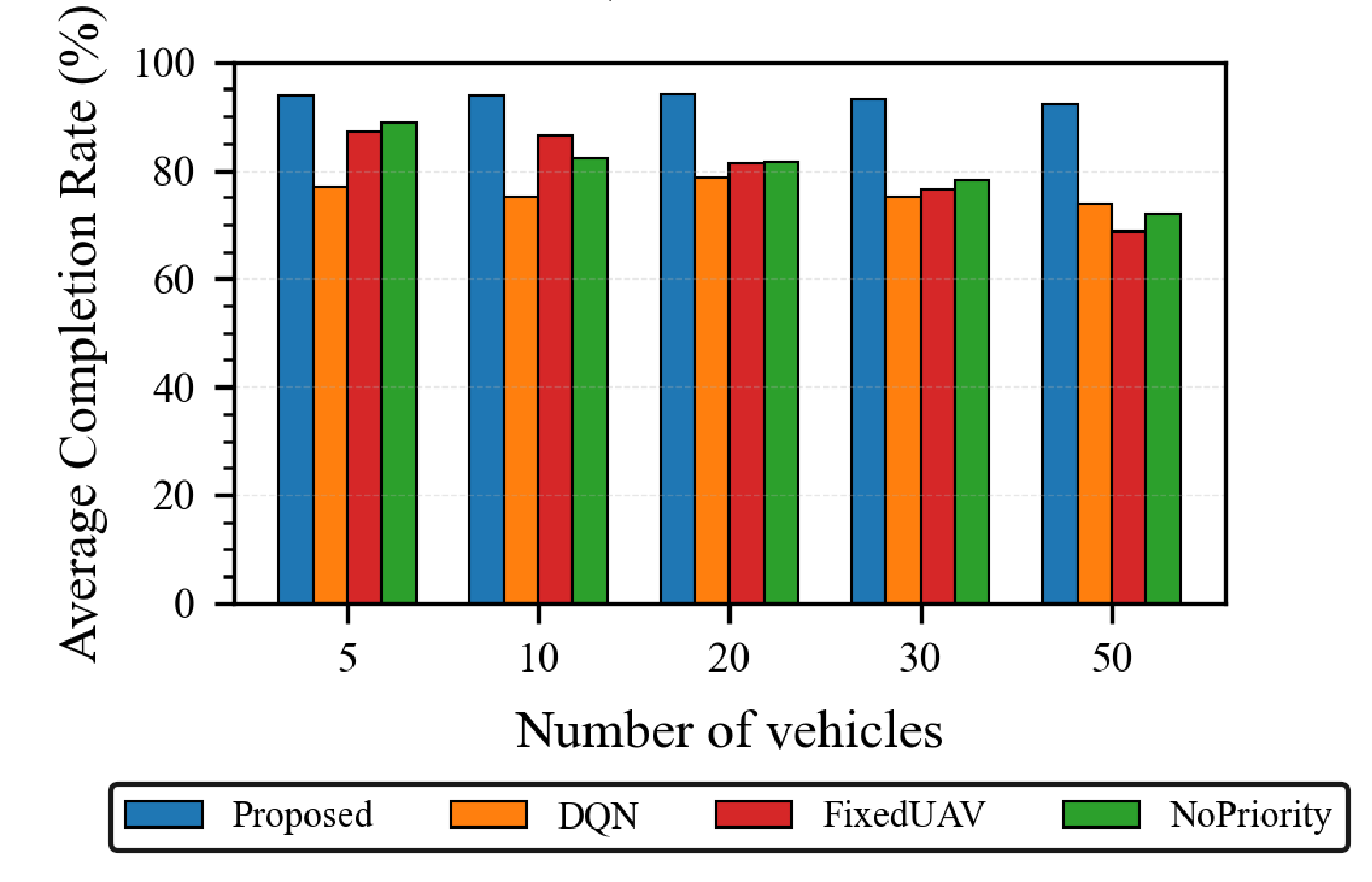}
    \caption{Task completion rate comparison under varying vehicle densities (LUAVs = 4, HUAV bandwidth = 100\,MHz).}
    \label{fig:performance}
\end{minipage}
\end{figure}

Fig.~\ref{fig:performance} illustrates the task completion rates of various offloading algorithms under different vehicle density conditions. The results demonstrate that the proposed algorithm consistently achieves a high completion rate as the number of vehicles scales from 5 to 50, with minimal performance fluctuations, which indicates strong robustness and adaptability to the network expansion. In contrast, all baseline algorithms exhibit noticeable performance degradation as vehicle density increases, particularly under medium to high load conditions, highlighting their limited stabilities in handling complex network scenarios. Overall, the proposed method effectively sustains task execution efficiency across diverse scales, validating its scalability and resilience in dynamic vehicular environments.
\begin{figure}[t]
    \begin{minipage}{0.48\textwidth}
    \centering
    \includegraphics[width=0.7\linewidth]{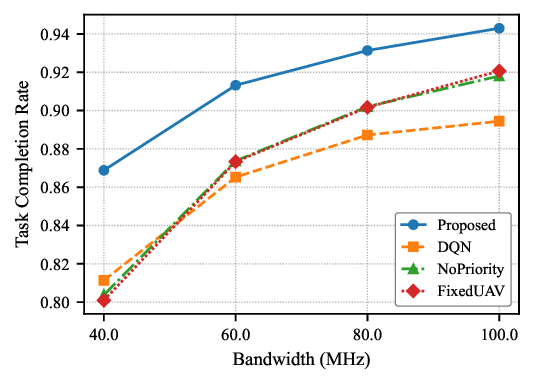}
    \caption{Task completion rate versus bandwidth allocation (20 vehicles, 4 LUAVs).}
    \label{fig:bandwidth}
\end{minipage}
\end{figure}

Fig.~\ref{fig:bandwidth} illustrates the impact of the HUAV bandwidth on task completion rates. The proposed method consistently outperforms other schemes, achieving 0.950 completion rate at 100\,MHz and maintaining stable performance of 0.875 at 40\,MHz. In contrast, FixedUAV shows the lowest performance of 0.820 at 40\,MHz, underscoring the limitations of static deployments under bandwidth constraints. The results highlight the importance of dynamic task offloading and LUAV trajectory scheduling, especially in low-bandwidth environments, where the flexible resource coordination becomes critical to maintain the system efficiency.

\section{Conclusions\label{sec:Conclusions}}
The paper proposes a dual-layer UAV-assisted computing architecture that integrates both aerial and ground resources. LUAVs provide low-delay edge offloading services, while a HUAV acts as a relay and coordination node, enhancing the system-wide computing cooperation and resource coverage. Based on the framework, the hierarchical SAC-based offloading algorithm is designed, which jointly considers task priority, computational load, and communication link quality, to enable efficient resource allocation and intelligent decision-making. Simulation results demonstrate that the proposed method outperforms existing approaches in terms of task completion rate, system utility, and convergence speed, exhibiting strong robustness and adaptability to complex vehicular environments.

\textcolor{black}
 {
     \bibliographystyle{IEEEtran}
     \bibliography{02ref/ref}

\begin{thebibliography}{10}
\providecommand{\url}[1]{#1}
\csname url@samestyle\endcsname
\providecommand{\newblock}{\relax}
\providecommand{\bibinfo}[2]{#2}
\providecommand{\BIBentrySTDinterwordspacing}{\spaceskip=0pt\relax}
\providecommand{\BIBentryALTinterwordstretchfactor}{4}
\providecommand{\BIBentryALTinterwordspacing}{\spaceskip=\fontdimen2\font plus
\BIBentryALTinterwordstretchfactor\fontdimen3\font minus \fontdimen4\font\relax}
\providecommand{\BIBforeignlanguage}[2]{{%
\expandafter\ifx\csname l@#1\endcsname\relax
\typeout{** WARNING: IEEEtran.bst: No hyphenation pattern has been}%
\typeout{** loaded for the language `#1'. Using the pattern for}%
\typeout{** the default language instead.}%
\else
\language=\csname l@#1\endcsname
\fi
#2}}
\providecommand{\BIBdecl}{\relax}
\BIBdecl

\bibitem{hu2020regional}
J.~Hu, C.~Chen, T.~Qiu, and Q.~Pei, ``Regional-centralized content dissemination for {eV2X} services in {5G} mmwave-enabled {IoV},'' \emph{IEEE Internet of Things Journal}, vol.~7, no.~8, pp. 7234--7249, Aug. 2020.

\bibitem{zhang2018vehicular}
S.~Zhang, J.~Chen, F.~Lyu, N.~Cheng, W.~Shi, and X.~Shen, ``Vehicular communication networks in the automated driving era,'' \emph{IEEE Commun. Mag.}, vol.~56, no.~9, pp. 26--32, Sep. 2018.

\bibitem{You2025CISS}
J.~You, Z.~Jia, C.~Dong, Q.~Wu, and Z.~Han, ``Generative {AI}-enhanced cooperative {MEC} of {UAVs} and ground stations for unmanned surface vehicles,'' in \emph{Proc. 59th Annu. Conf. Inf. Sci. Syst. (CISS)}, Mar. 2025.

\bibitem{liu2023uav}
R.~Liu, A.~Liu, Z.~Qu, and N.~N. Xiong, ``An {UAV}-enabled intelligent connected transportation system with {6G} communications for {I}nternet of vehicles,'' \emph{IEEE Trans. Intell. Transp. Syst.}, vol.~24, no.~2, pp. 2045--2059, Feb. 2023.

\bibitem{hu2021uav}
J.~Hu, C.~Chen, L.~Cai, M.~R. Khosravi, Q.~Pei, and S.~Wan, ``{UAV}-assisted vehicular edge computing for the {6G} {I}nternet of vehicles: Architecture, intelligence, and challenges,'' \emph{IEEE Commun. Stand. Mag.}, vol.~5, no.~2, pp. 12--18, Jun. 2021.

\bibitem{Jia2024VTM}
Z.~Jia, J.~You, C.~Dong, Q.~Wu, F.~Zhou, D.~Niyato, and Z.~Han, ``Cooperative cognitive dynamic system in {UAV} swarms: Reconfigurable mechanism and framework,'' \emph{IEEE Veh. Technol. Mag.}, vol.~19, no.~3, pp. 90--101, Sep. 2024.

\bibitem{10638237}
Y.~Wu, Z.~Jia, Q.~Wu, and Z.~Lu, ``Adaptive qoe-aware sfc orchestration in uav networks: A deep reinforcement learning approach,'' \emph{IEEE Trans. Netw. Sci. Eng.}, vol.~11, no.~6, pp. 6052--6065, Nov.-Dec. 2024.

\bibitem{jia2025dro}
Z.~Jia, C.~Cui, C.~Dong, Q.~Wu, Z.~Ling, D.~Niyato, and Z.~Han, ``Distributionally robust optimization for aerial multi-access edge computing via cooperation of {UAVs} and {HAPs},'' \emph{IEEE Trans. Mob. Comput.}, vol.~--, no.~--, pp. 1--15, 2025.

\bibitem{li2023flexedge}
B.~Li, W.~Xie, Y.~Ye, L.~Liu, and Z.~Fei, ``Flexedge: Digital twin-enabled task offloading for {UAV}-aided vehicular edge computing,'' \emph{IEEE Trans. Veh. Technol.}, vol.~72, no.~8, pp. 11\,086--11\,091, Aug. 2023.

\bibitem{yu2020joint}
Z.~Yu, Y.~Gong, S.~Gong, and Y.~Guo, ``Joint task offloading and resource allocation in {UAV}-enabled mobile edge computing,'' \emph{IEEE Internet Things J.}, vol.~7, no.~4, pp. 3147--3159, Apr. 2020.

\bibitem{tun2021energy}
Y.~K. Tun, Y.~M. Park, N.~H. Tran, W.~Saad, S.~R. Pandey, and C.~S. Hong, ``Energy-efficient resource management in {UAV}-assisted mobile edge computing,'' \emph{IEEE Commun. Lett.}, vol.~25, no.~1, pp. 249--253, Jan. 2021.

\bibitem{jia2025sfc}
Z.~Jia, Y.~Cao, L.~He, Q.~Wu, Q.~Zhu, D.~Niyato, and Z.~Han, ``Service function chain dynamic scheduling in space-air-ground integrated networks,'' \emph{IEEE Trans. Veh. Technol.}, vol.~--, no.~--, pp. 1--15, 2025.

\bibitem{dai2024uav}
X.~Dai, Z.~Xiao, H.~Jiang, and J.~C.~S. Lui, ``{UAV}-assisted task offloading in vehicular edge computing networks,'' \emph{IEEE Trans. Mobile Comput.}, vol.~23, no.~4, pp. 2520--2534, Apr. 2024.

\bibitem{9026875}
H.~Peng and X.~Shen, ``Deep reinforcement learning based resource management for multi-access edge computing in vehicular networks,'' \emph{IEEE Transactions on Network Science and Engineering}, vol.~7, no.~4, pp. 2416--2428, Oct.-Dec. 2020.

\bibitem{Diao2021UAV}
X.~Diao, W.~Yang, L.~Yang, and Y.~Cai, ``{UAV}-relaying-assisted multi-access edge computing with multi-antenna base station: Offloading and scheduling optimization,'' \emph{IEEE Trans. Veh. Technol.}, vol.~70, no.~9, pp. 9495--9509, Sep. 2021.

\bibitem{10616106}
K.~Mao, Q.~Zhu, C.-X. Wang, X.~Ye, J.~Gomez-Ponce, X.~Cai, Y.~Miao, Z.~Cui, Q.~Wu, and W.~Fan, ``A survey on channel sounding technologies and measurements for {UAV}-assisted communications,'' \emph{IEEE Trans. Instrum. Meas.}, vol.~73, pp. 1--24, Jan. 2024.

\bibitem{wang2024sparse}
J.~Wang, Q.~Zhu, Z.~Lin, J.~Chen, G.~Ding, Q.~Wu, G.~Gu, and Q.~Gao, ``Sparse bayesian learning-based hierarchical construction for 3d radio environment maps incorporating channel shadowing,'' \emph{IEEE Trans. Wireless Commun.}, vol.~23, no.~10, pp. 14\,560--14\,574, 2024.

\bibitem{9714482}
Z.~Jia, Q.~Wu, C.~Dong, C.~Yuen, and Z.~Han, ``Hierarchical aerial computing for {I}nternet of things via cooperation of {HAPs} and {UAVs},'' \emph{IEEE Internet Things J.}, vol.~10, no.~7, pp. 5676--5688, Apr. 2023.

\end{thebibliography}
 }
\end{document}